\documentclass[10pt,twocolumn,letterpaper]{article}
\usepackage{cvpr}
\usepackage{times}
\usepackage{epsfig}
\usepackage{graphicx}
\usepackage{amsmath}
\usepackage{amssymb}
\usepackage{algorithm}  
\usepackage{algpseudocode}  
\usepackage{booktabs}
\usepackage{bbding} 
\usepackage[pagebackref=true,breaklinks=true,colorlinks,bookmarks=false]{hyperref}

\cvprfinalcopy % *** Uncomment this line for the final submission
\def\cvprPaperID{2604} % *** Enter the CVPR Paper ID here

\ifcvprfinal\fi
\begin{document}

%%%%%%%%% TITLE
\title{PVT: Point-Voxel Transformer for Point Cloud Learning}

\author{Cheng Zhang$^1\thanks{These authors contributed equally.}$, Haocheng Wan$^{1*}$, Xinyi Shen$^2$, Zizhao Wu$^{1}\thanks{Corresponding author.}$\\
$^1$Hangzhou Dianzi University, Hangzhou China\\
$^2$University College London, London UK\\
{\tt\small \{zhangcheng828,wanhaocheng2022,xinyishen2018,wuzizhao\}@foxmail.com, @163.com, @hdu.edu.cn}}
% \and
% Haocheng Wan\\
% Hangzhou Dianzi University\\
% Hangzhou China\\
% {\tt\small wanhaocheng2022@163.com}
% \and
% Xinyi Shen\\
% University College London\\
% London UK\\
% {\tt\small xinyishen2018@163.com}
% }

\maketitle
%\thispagestyle{empty}

%%%%%%%%% ABSTRACT
\begin{abstract}
The recently developed pure Transformer architectures have
attained promising accuracy on point cloud learning benchmarks compared to convolutional neural networks. However, existing point cloud Transformers are computationally expensive because they waste
a significant amount of time on structuring irregular data. To solve this shortcoming, we present the Sparse Window Attention (SWA) module to gather coarse-grained local features from non-empty \emph{voxels}. The module not only bypasses the expensive irregular data structuring and invalid empty voxel computation, but also obtains linear computational complexity with respect to voxel resolution. Meanwhile, we leverage two different self-attention variants to gather fine-grained features about the global shape according to different scale of point clouds.
Finally, we construct our neural architecture called Point-Voxel Transformer (PVT), which integrates these modules into a joint framework for point cloud learning. Compared with previous Transformer-based and attention-based models, our method attains a top accuracy of 94.1\% on the classification benchmark and 10$\times$ inference speedup on average. Extensive experiments also validate the effectiveness of PVT on semantic segmentation benchmarks. Our code and pretrained model are avaliable at \url{https://github.com/HaochengWan/PVT}.
\end{abstract}


%%%%%%%%% BODY TEXT
\section{Introduction}
Point cloud learning has been receiving increasing attention from both industry and academia due to its wide applications including autonomous driving, robotics, AR/VR, etc. In these settings, sensors like LIDAR produce irregular and unordered sets of points that correspond to object surfaces. However, how to capture semantics directly and quickly from these data remains a challenge for point cloud learning.

Most existing point cloud learning methods can be classified into two categories in terms of data representations: \emph{voxel}-based models and \emph{point}-based models. The \emph{Voxel}-based models generally rasterize point clouds onto regular grids and apply 3D convolution for feature learning \cite{zhou2018voxelnet,2016Volumetric,2019VoxSegNet}. These models are computationally efficient due to their excellent memory locality, but the inevitable information loss degrades the fine-grained localization accuracy \cite{2019PV}. By contrast, \emph{point}-based models naturally preserve the accuracy of point location, but are generally computationally intensive
\cite{Spectral,2020ClusterNet,2019Modeling}.

Recently, Transformer architecture has drawn great attention in natural
language processing and 2D vision because of its superior capability in capturing long-range dependencies.
Powered by Transformer \cite{2017Attention} and its variants \cite{liu2021swin,DBLP:journals/corr/abs-2010-11929}, \emph{point}-based models have applied self-attention (the core unit of Transformer) to extract features from point clouds and improve performance significantly \cite{guo2020pct,Nico,zhao2020point,2020PointASNL}. However, most of them suffer from the time-consuming process of sampling and aggregating features from irregular points, which becomes the efficiency bottleneck \cite{2019Point}.

In this paper, we study how to design an efficient and high-performance Transformer architecture while avoiding the shortcoming of previous point cloud Transformers. We observe that \emph{voxel}-based models have regular data locality and can efficiently encode coarse-grained features, while \emph{point}-based networks preserve the accuracy of location information with the flexible fields and can effectively aggregate fine-grained features. Inspired by this, we propose a novel point cloud learning architecture, namely, Point-Voxel Transformer (PVT), which combines the ideas of \emph{voxel}-based and \emph{point}-based models. Our network focuses on how to fully exploit the potential of the two models mentioned above in Transformer architecture, capturing useful discriminative features from 3D data.

To investigate this, we conduct self-attention (SA) computation in \emph{voxels} to obtain efficient learning pattern while performing SA in \emph{points} to preserve the accuracy of location information with the flexible fields. 
However, directly performing SA computation to voxels is infeasible, mainly owing to two facts. First, non-empty voxels are sparsely distributed in the \emph{voxel} domain and only account for a small proportion of total voxels \cite{SECOND}. Second, the original SA computation leads to quadratic complexity with respect to the number of voxels, making it unsuitable for various 3D vision tasks, particularly for the dense prediction tasks of object detection and semantic segmentation.
To tackle these issues, we design a sparse window attention (SWA) module which only has linear computational complexity by computing SA locally within the non-overlapping 3D window.
A key design element of SWA lies in its sparse attention computing, in which
a GPU-based Rule Book stores the non-empty voxel indexes. On the basis of this strategy, we can bypass the invalid computation of empty voxels and retain the original 3D shape structure. For cross-window information interaction, inspired by Swin Transformer \cite{liu2021swin}, we propose to apply shifted window to enlarge the receptive field. In addition, we leverage two different SA variants to capture the global information according to different scales of point clouds. For small-scale point clouds, we propose relative-attention (RA), which extends the SA mechanism to consider representations of the relative position, or distances between points. The advantage of RA is that the absolute coordinates of the same object can be completely different with rigid transformations; thus, injecting relative position representations (RPR) in our structure is generally more robust.
We observe that our network obtains significant improvements over not using this RPR term without adding extra training parameters. Nevertheless, with tens of thousands of points (e.g., SemanticKITTI \cite{semanticKT}) as inputs, directly applying the RA module in \emph{points} incurs unacceptable $\mathcal{O}(N^2)$ memory consumption, where $N$ is the input point number. Thus, for large-scale point clouds, we perform External Attention (EA), a linear attention variant to avoid the $\mathcal{O}(N^2)$ computational complexity of RA module.

Based on these modules, We propose a two-branch PVT that mainly consists of the \emph{voxel} branch and the \emph{point} branch (Figure \ref{PVTLayer}). As illustrated in Figure \ref{fig:disentangle}, by behaving SWA in the \emph{voxel} branch and performing RA (or EA) in the \emph{point} branch, our method disentangles the coarse-grained local feature aggregation and the fine-grained global context transformation so that each branch can be implemented efficiently and accurately.

By using the PVT, we construct PVT networks for various 3D tasks. These networks can capably serve as a general-purpose backbone for point cloud learning. In particular, we conduct the classiﬁcation experiment on the ModelNet40 \cite{modelnet40} dataset and obtain the state-of-the-art accuracy of 94.1\% (no voting), while being on average 10$\times$ faster than its Transformer baselines. On ShapeNet Part \cite{shapenet}, S3DIS \cite{DBLP:journals/corr/ArmeniSZS17}, and SemanticKITTI \cite{semanticKT} datasets, our model also obtains strong performance (86.6\%, 69.2\%, and 64.9\% mIoU, respectively). 

The main contributions are summarized as following:
\begin{itemize}
\item We propose PVT, the first Transformer-based approach, to deeply incorporate the advantages from both \emph{point}-based and \emph{voxel}-based networks to our knowledge.
\item We propose an efficient local attention module, named SWA, which achieves linear computational complexity with respect to the input voxel size and bypasses the invalid empty voxel computation.
\item Extensive experiments show that our PVT model attains competitive results on general point cloud learning tasks with 10$\times$ latency speed-up than its baselines.
\end{itemize}
%------------------------------------------------------------------------
\section{Related works}

\subsection{Voxel-based models on points}
To leverage the success of convolutional neural networks on 2D images, many researchers have endeavored to project 3D point clouds directly onto voxels and perform 3D convolution for information aggregation \cite{20194D,2018PointGrid}. However, the memory and computational consumption of such full-voxel-based models increases cubically with respect to the voxel's resolution. To tackle this shortcoming, O-CNN \cite{2017O}, OctNet \cite{2016OctNet}, and kd-Net \cite{klokov2017escape} constructed tree structures for 3D voxels to bypass the invalid computation of the empty voxels. Although such methods are efficient in data structuring, geometric details are lost during the projection because multiple bucketing points into the same voxel leads to indistinguishable features for learning.

Compared with most \emph{voxel}-based 3D models, our model is more efficient and effective for the following reasons: 1) Instead of using 3D convolutions, we propose SWA to handle the sparsity characteristic of voxels that have a linear computational complexity with respect to voxel resolution and bypasses the invalid computation of empty voxels. 2) As we employ RA in the \emph{point} domain, an inherent advantage is that we can avoid feature loss during voxelization, maintaining the geometric details of a point cloud.
\subsection{Point-based models for point cloud learning}

Instead of voxelization, it is possible to make a neutral network that consumes directly on point clouds. \cite{qi2017pointnet} proposed PointNet, the pioneering work that learns directly on sparse and unstructured point clouds.
Inspired by PointNet, previous works introduced sophisticated neural modules to learn per-point features. These models can be generally classified as: 1) neighbouring feature pooling \cite{2019PointWeb,2018Recurrent}, 2) graph message passing \cite{2018Dynamic,2019Linked,2019Graph}, and 3) attention-based or Transformer-based models \cite{2021Semantic,2019PCAN,2019Learning,guo2020pct,zhao2020point,Nico}. 
\begin{figure*}
    \centering
    \includegraphics[width=\linewidth]{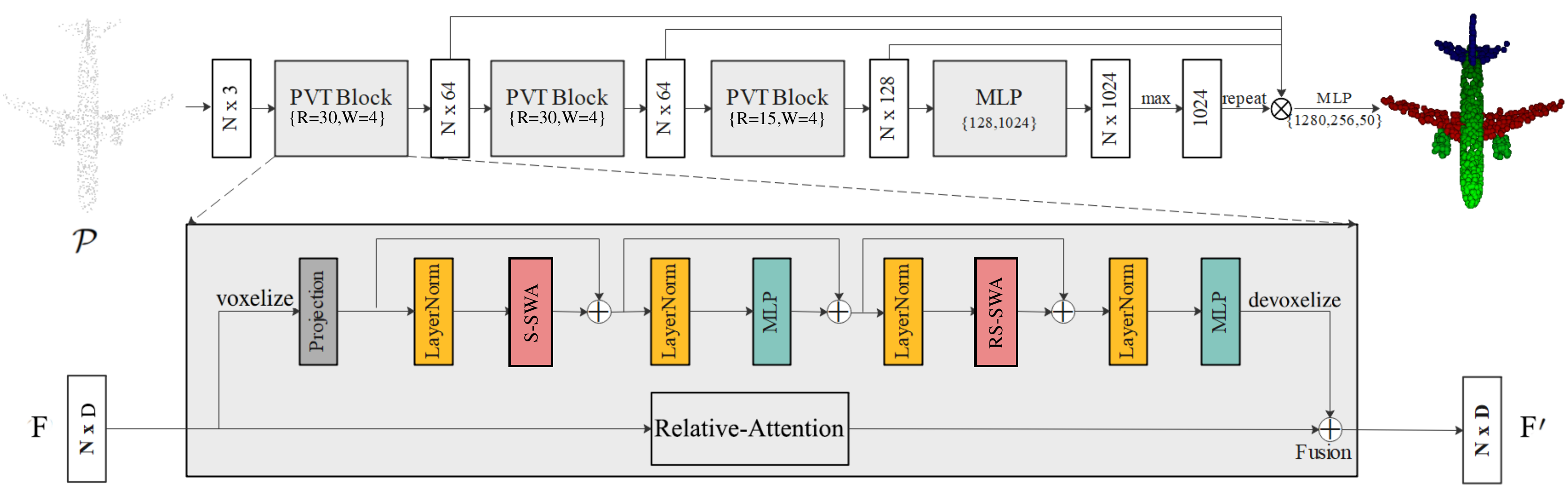}
    \caption{\textbf{Model Architecture: }The model for part segmentation take as input N points and feeds them into 3 stacked PVT blocks to learn a semantically rich and discriminative representation for each point, followed by a MLP layer to generate the output feature. After that, we leverage the max-pooling and repeating operators to extract an effective global feature representing the entire point cloud. Note that shortcut connections are used to extract multi-scale features and one MLP layer (1280) to aggregate multi-scale features, where we concatenate features from previous layers to get a 64+64+128+1024=1280-dimensional point cloud. Finally, we predict the final point-wise segmentation scores for the input point cloud and the part label of a point is also determined as the one with maximal score.
    \textbf{PVT block (grey box):} The PVT block is composed of two branches. The upper branch is \emph{voxel}-based for aggregating local features and the lower is \emph{point}-based for capturing global features. We can effectively fuse two branches because they are providing complementary information. $R$ and $w$ denote the voxel resolution size and 3d window size, respectively. $\bigoplus$: addition, $\bigotimes$: concatenation.}
    \label{PVTLayer}
\end{figure*} 

Owing to the sparsity and irregularity of point clouds, the methods that directly consume points have achieved state-of-the-art performance. However, the cost of data structuring of such methods has become the computation burden, especially on the large-scale point clouds dataset \cite{2019Point,xu2020grid}. In this study, we handle this shortcoming by using the SWA module.

\subsection{Self-attention and Transformer}
\cite{bahdanau2014neural} proposed a neural machine translation method with an attention mechanism, in which attention weight is computed through the hidden state of an RNN. Then \cite{lin2017structured} further proposed self-attention to visualize and interpret sentence embeddings. 
%In their work, the self-attention first takes the sum of input data and positional embedding as input, and computes three vectors for each word: $q$, $k$ and $v$ through trained linear layers. Then, the attention weight between any two words is obtained by dot-producting $q$ and $k$ vectors. Finally, the attention feature is defined as the weighted sum of all $v$ vectors with the attention weights. %
Subsequent works employed self-attention layers to replace some or all the spatial convolution layers, such as Transformer for machine translation \cite{2017Attention}. \cite{2018BERT} proposed the bidirectional transformers (BERT), which is one of the most powerful models in the NLP field. 

Given the success of self-attention and Transformer architectures in NLP, researchers have applied
them to vision tasks \cite{2019Local,DBLP:journals/corr/abs-1906-05909,2020Exploring}. For instance, \cite{DBLP:journals/corr/abs-2010-11929} proposed an image recognition network, ViT, which directly applied a Transformer architecture on image patches and achieved better performance than the traditional convolutional neural networks. \cite{liu2021swin} recently introduced Swin Transformer to incorporate inductive bias for spatial locality, hierarchy and translation invariance.

\subsection{Transformers on point cloud}
Recently, plenty of researchers have attempted to explore Transformer-based architectures for point cloud learning. \cite{Nico} proposed PT$^1$ to extract global features by introducing the dot-product SA mechanism. \cite{guo2020pct} proposed offset-attention to calculate the offset difference between the SA features and the input features by element-wise subtraction. Moreover, \cite{2020Point} proposed PT$^2$ to build local vector attention in neighborhood point sets and achieved significant progress. However, they wasted a high percentage of the total time on structuring the irregular data, which becomes the efficiency bottleneck. In this work, we study how to solve this shortcoming, so as to design an efficient Transformer-based architecture for point cloud analysis.

There also exists some Transformer-based architecture for various point cloud processing tasks. For instance, \cite{se3} proposed the SE(3)-Transformer, a variant of the self-attention module for 3D point clouds and graphs, which is equivariant under continuous 3D roto-translations. Late, \cite{pointtr} and \cite{snow} have attempted to explore Transformer-based architectures for point cloud completion and achieved significant performance on all completion tasks. Recently, \cite{point-bert} proposed Point-BERT to unleash the scalability and generalization of Transformers for 3D point cloud representation learning. Extensive experiments also demonstrate that Point-BERT significantly improves the performance of standard point cloud Transformers. Different from these Transformers, in this paper, we study how to design an efficient and high-performance Transformer architecture for point cloud learning, making it suitable for edge devices with limited computational resources or real-time autonomous driving scenarios.
\section{Method}

\noindent \textbf{Overview}

In contrast to the previous point cloud Transformers \cite{guo2020pct,Nico,zhao2020point}, we design the network as efficient as possible with high accuracy so that it can be widely used on various 3D tasks.

We introduce an efficient neural architecture for point cloud learning, namely, PVT. Formally, given a point cloud embedding $F \in \mathbb{R}^{N\times D}$, our PVT is designed to map the input features $F$ to a new set of point feature $F^{\prime} \in \mathbb{R}^{N\times D}$. As illustrated in Figure \ref{PVTLayer}, the PVT consists of two main branches: a \emph{voxel} branch and a \emph{point} branch. We leverage the \emph{voxel} branch to map the inputs $F$ to $F_{local}\in \mathbb{R}^{N\times D}$, which aggregates local features in the \emph{voxel} domain. However, full \emph{voxel} method will inevitably encounter information loss during voxelization. Thus, we utilize the \emph{point} branch to map the inputs $F$ to $F_{global}\in \mathbb{R}^{N\times D}$, which can directly extract global features for each individual point. With both local features and aggregated global context, we can efficiently fuse two branches into an addition layer as both provide complementary information.

Below we detail the two branches in Sections \ref{section3.1} and \ref{3.2}. Section \ref{3.3} details our feature fusion module, and Section \ref{3.4} discusses the relationship between the proposed model and the prior works.

\subsection{Voxel branch} \label{section3.1}
This branch aims to effectively capture local information, which can bypass expensive sampling and neighbor points querying. Specifically, it contains three steps: voxelization, feature aggregation, and devoxelization.

The voxelized and devoxelized methods map the input point cloud $\mathcal{P}$ to a new set of voxel features $V$ and transform the voxel-wise features back to the point-wise features $F_{local}$ (Readers can refer to \cite{2019Point} for more details). In this work, we apply the standard Transformer architecture \cite{2017Attention} to perform feature aggregation on regular 3D voxels, which can improve accuracy significantly. However, the standard Transformer architecture conducts global-SA which leads to quadratic complexity with respect to the number of voxels, making it unsuitable for many 3D vision problems requiring dense prediction, such as semantic segmentation.

\textbf{Window Attention:} To obtain efficient modeling power, we propose to compute SA within local 3D windows, which are arranged to evenly partition the voxel space in a non-overlapping manner. Assume that each local 3D window contains $W\times W\times W$ voxels and $R$ denotes the voxel resolution. The computational complexity of a global-SA and a window-based one on $R^3$ voxels are:
\begin{alignat}{1}
    \Omega(Global\textbf{-}SA)=4R^3D^2+2(R^3)^2D \\
    \Omega(Window\textbf{-}SA)=4R^3D^2+2B^3R^3D
\end{alignat}
where the former is quadratic to the number of voxels $R^3$, the latter is linear when $W$ is fixed, $D$ is the dimension of features. In summary, global-SA computation is generally unaffordable for a large voxel resolution, whereas the local window-SA is scalable.

\begin{figure}
    \centering
    \includegraphics[width=\linewidth]{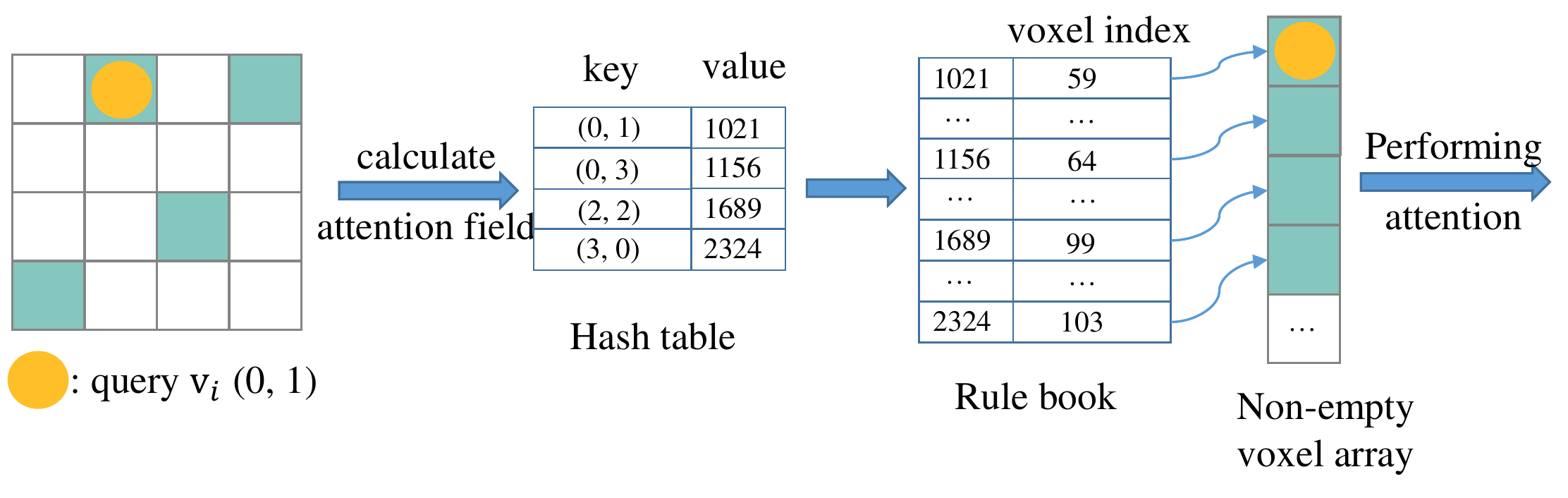}
    \caption{\textbf{Sparse Window Attention: }This is also a 2D example and can be easily extended to 3D cases. SWA is a GPU-based method and can efficiently index the non-empty voxels in the attention field.}
    \label{SWA}
\end{figure}

\textbf{Sparse Window Attention:} Different from 2D pixels
densely placed on an image plane, non-empty voxels only account for a small proportion of total voxels. Inspired by \cite{sparse,SECOND}, we design Sparse Window Attention to handle the sparsity characteristic of voxels. As shown in Figure \ref{SWA}, for each querying index $v_i$, we first use Window Attention to determine all neighboring voxel indices in the attention field, and then adopt a Hash table to get the hashed integer neighboring voxel indices. Last, a GPU-based Rule Book stores the hashed voxel indices as keys, and the corresponding indices for the non-empty voxel array as values. Finally, we can perform Window Attention to gather the coarse-grained local features. Note that all the steps can be conducted in parallel on GPUs by assigning each querying voxel $v_i$ a separate CUDA thread.

The SWA lacks information interaction across windows, which may limit the representation power of our model.
Thus, we extend the shifted 2D window mechanism of Swin Transformer \cite{liu2021swin} to 3D windows for the purpose of introducing cross-window information interaction while maintaining the efficient computation of non-overlapping SWA. Readers can refer to Swin Transformer for more details

As shown in Figure \ref{PVTLayer}, with the cyclic shifted window partitioning approach, the step of feature aggregation of this branch can be described as follows:

\begin{algorithm}[H]
    \caption{ Pseudo-code for the Voxel Branch}%标题
    \label{Pseudo-code}%标签
    \hspace*{0.02in} { \textbf{Input:} $P$, an array with shape [B, N, C]. $F$, an array with shape [B, N, D].} \\
    \hspace*{0.02in} { \textbf{Output:} $F_{local}$, an array with shape [B, N, D]}\\
    \hspace*{0.04in} F\_1 = voxelize(P, F) $\textcolor{cyan}{\# shape=[B,R^3,D]}$\\
    \hspace*{0.04in} F\_1 = cyclic\_shift(F\_1)\\
    \hspace*{0.04in} F\_2 = SWA(layernorm(F\_1)) + F\_1\\
    \hspace*{0.04in} F\_2 = MLP(layernorm(F\_2)) + F\_2\\
    \hspace*{0.04in} F\_2 = reverse\_cyclic\_shift (F\_2)\\
    \hspace*{0.04in} F\_3 = SWA(layernorm(F\_2)) + F\_2\\
    \hspace*{0.04in} $F_{local}$ = MLP(layernorm(F\_3)) + F\_3\\
\end{algorithm}

\subsection{Point branch}\label{3.2}
The \emph{voxel} branch gathers the neighborhood information with low resolution. However, in order to capture long-range dependencies, low-resolution \emph{voxel} branch alone is limited. To this end, we attempt to employ the following self-attention on the entire point cloud for global context aggregation, which is computed as:
\begin{alignat}{1}
F_{sa} &= softmax(QK^T) V, \quad F_{sa}\in \mathbb{R}^{N\times D} \label{attn}\\
F_{global} &= MLP(F_{sa})+F,\quad F_{global}\in \mathbb{R}^{N\times D} 
\end{alignat}
where (Q, K, V) $\in \mathbb{R}^{N\times D}$ is generated by shared linear transformations and the input features $F$ and are all ordered independent. Moreover, \emph{softmax} and \emph{weighted sum} are both permutation-independent operators. Thus, the self-attention computation is permutation-invariant, making it well-suited to handle the irregular, disordered 3D points.

\textbf{Relative Attention:} Self-attention mentioned above fails to incorporate relative position representations in its structure, whereas such ability is very important to 3D visual tasks. For example, the absolute coordinates of the same object can be completely different with rigid transformations. Therefore, injecting relative position representations are generally more robust.
In this study, we design our relative-attention to embed relative position representations, which are not well studied in prior point cloud learning works.

First of all, by embedding RPR into the scaled dot-product self-attention module, Eq \ref{attn} can be re-formulated as:
\begin{align}
    F_{ra} = softmax(QK^{T}+B)V
\label{bias}
\end{align}
where $B\in \mathbb{R}^{N\times N}$ is the relative representations bias. 

Suppose that the original point cloud with $N$ points is denoted by $\mathcal{P}= \{p_i\}_{i=1}^N \subseteq \mathbb{R}^3 $. We compute the relative position $B$ as follows:
\begin{align}
    B_{p_i,p_j,m} = p_{i,m}-p_{j,m}, \quad m \in \{x,y,z\}.
\end{align}

To map relative coordinates to the corresponding position encoding, we maintain three learnable look-up tables $t_x, t_y, t_z \in \mathbb{R}^{L\times N\times N}$ corresponding to the x, y and z axis, respectively. As the relative coordinates are continuous floating-point numbers, we uniformly quantize the range of $B_{p_i,p_j,m}$ into $L$ discrete parts and map the relative coordinates $B_{p_i,p_j,m}$ to the indices of the tables as:
\begin{align}
    idx_{i,j,m} = \lfloor \frac{B_{p_i,p_j,m} + s_{max}} {s_{quad}} \rfloor
\end{align}
where $s_{max}$ is the maximum size of point cloud coordinates and $s_{quad} = \frac{2\cdot s_{max}}{L}$ is the quantization size, and $\lfloor \cdot \rfloor$ denotes floor rounding.

We look up the tables to retrieve corresponding embedding with the index and sum them up to obtain the position encoding of
\begin{align}
    B_{p_i,p_j} = \sum_{m=1}^3 t_m[idx_{i,j,m}],
\end{align}
where $t[idx]$ indicates the $idx$-th entry of the learnable look-up table $t$, and $B_{p_i,p_j}$ means the relative position encoding between $p_i$ and $p_j$.

\textbf{External Attention:} 
Although the RA module has demonstrated significant performance on small-scale point clouds, it is unsuitable for large-scale point clouds (e.g., SemanticKITTI \cite{semanticKT}) due to its unacceptable $\mathcal{O}(N^2)$ memory consumption. Therefore, in this work, we perform External Attention computation on large-scale point clouds. External Attention, is a novel attention mechanism based on two external, small, learnable, shared memories, which can be implemented easily by simply using two cascaded linear layers and two normalization layers. It has linear complexity and implicitly considers the correlations between all data samples, making it suitable for large-scale point clouds. 

\subsection{Feature fusion} \label{3.3}
We effectively fuse the outputs of two branches with an addition as they are providing complementary information: 
\begin{alignat}{1}
F^{\prime} = F_{local} + F_{global},\quad
F^{\prime} \in \mathbb{R}^{N\times D}
\end{alignat}

\subsection{Relationship to prior works} \label{3.4} 
\begin{table}
\centering
\begin{center} 
    \begin{tabular}{|l|lc|}
    \hline
        & Layer Type      & Time Complexity per Layer  \\
    \hline
       The- & 3D Convolutions & $\mathcal{O}(k\cdot R^3\cdot D^2)$ \\
       voxel-  & Window-attention &     $\mathcal{O}(v\cdot R^3\cdot D)$ \\
       branch  & SWA &     $\mathcal{O}(r^2\cdot v\cdot R^3\cdot D)$ \\
       \hline
       The- & 1D Convolutions &  $\mathcal{O}(k\cdot N\cdot D^2)$ \\
       point- & Relative-attention &   $\mathcal{O}(N^2\cdot D)$ \\
       branch & External-attention &   $\mathcal{O}(N\cdot D)$ \\
  \hline
    \end{tabular}
    \end{center}
     \caption{Per-layer complexity for different layer types. $R$ is the resolution of voxels, $v = W\times W\times W$ is the number of voxels in a same 3D window, $r$ denotes the proportion of non-empty voxels in a local 3D window, $N$ is the number of points, $D$ is the representation dimension, and $k$ is the kernel size of convolutions.}
    \label{Complexity}
    
\end{table}

The proposed PVT is related to several prior works which includes PVCNN \cite{2019Point}, PCT \cite{guo2020pct}, PT$^1$\cite{Nico}, PT$^2$\cite{zhao2020point},.

Although we are inspired by the idea of PVCNN \cite{2019Point}, our PVT is different in several ways: 1) in the \emph{voxel} branch, PVCNN uses a 3D convolution to gather local information while we employ SWA computation within each local window, which is highly efficient. Per-layer complexity for different layer types are shown in Table \ref{Complexity}, for large-scale point clouds, approximately 10\% of the voxels are non-empty ($r=0.1$) \cite{xu2020grid} which means that our SWA can save significant computation power compared with window-attention.
2) in the \emph{point} branch, PVCNN uses 1D convolution, which is computation efficient but lacks the global context modeling capability. By performing RA (or EA) computation on the entire points, our method gathers the global modeling power.

Unlike prior Transformer-based 3D models that need to gather the downsampled points and find the corresponding neighbors by using expensive FPS and $k$-NN in \emph{point} domain, our approach does not require explicitly identify which point is the farthest and what are in the neighboring set. Instead, the \emph{voxel} branch observes regular data and learns to capture local features using SWA. Additionally, the \emph{point} branch only needs to perform RA (or EA) on the entire point cloud, which also does not require to find the neighboring points. Thus, our approach obtains highly efficiency than PCT, PT$^1$ and PT$^2$ (see Table \ref{computational}).

\section{Experiments}
We now discuss the PVT that can be constructed from PVT blocks for different point cloud learning tasks: shape classification, and object and semantic segmentation. The performance is quantitatively evaluated with four metrics, including overall accuracy (OA), average precision (AP), the intersection over union (IoU), and mean IoU (mIoU). For the sake of fair comparison with baselines, we report the measured latency on a GTX 2080 GPU to reﬂect the efﬁciency but evaluate other indicators on a GTX 3090 GPU.

\textbf{Model.}
The architecture used for the segmentation task is shown in Figure \ref{PVTLayer}. In our settings, dropout with keep probability of 0.5 is used in the last two linear layers. All layers include ReLU and batch normalization. In addition, for other point cloud learning tasks, we use the similar architecture as in segmentation. Readers can refer to our source code for more derails.

\textbf{Baselines.}
We select four models as the comparison baselines, including the strong attention-based model PointASNL and the three powerful Transformer-based networks PT$^1$, PT$^2$, and PCT.

\subsection{Shape Classification}
\begin{table}  
\centering
\begin{center} 
\begin{tabular}{|l|cccc|}   
\hline  
Model & Input & Points & OA(\%) & Latency\\ 
\hline
\multicolumn{5}{|c|}{OA$<$92.5} \\
\hline
PointNet  & xyz & 16$\times$1024 &  89.2  & \textbf{13.6ms}\\ 
PointNet++  \cite{qi2017pointnet++} & xyz,nor & 16$\times$1024 &  91.9  & 35.3ms\\  
SO-Net \cite{2018SO}  & xyz,nor & 8$\times$2048 & 90.9   & $-$\\
PointGrid \cite{2018PointGrid}  & xyz,nor & 16$\times$1021 & 92.0   & $-$\\
SpiderCNN \cite{2018SpiderCNN}  & xyz,nor & 8$\times$1024 & 92.4   &82.6ms \\
PointCNN \cite{2018PointCNN} & xyz & 16$\times$1024 & 92.2  & 221.2ms\\  
PointWeb \cite{zhao2019pointweb}  & xyz,nor & 16$\times$1024 & 92.3  & $-$\\  
PVCNN \cite{2019Point} & xyz,nor & 16$\times$1024  & 92.4 & 24.2ms\\
\hline
\multicolumn{5}{|c|}{OA$>$92.5} \\
\hline
KPConv \cite{2019KPConv}  & xyz &16$\times$6500 & 92.9   &120.5ms \\ 
DGCNN \cite{2018Dynamic}  & xyz & 16$\times$1024 & 92.9  &  85.8ms\\
LDGCNN \cite{2019Linked}  & xyz & 16$\times$1024 & 92.7  &$-$ \\
PointASNL \cite{2020PointASNL} & xyz,nor & 16$\times$1024 & 93.2  & 923.6ms\\
PT$^1$ \cite{Nico}& xyz,nor & 16$\times$1024 & 92.8 & 320.6ms\\
PT$^2$ \cite{zhao2020point} & xyz,nor  & 8$\times$1024 & 93.7  & 530.2ms \\
PCT \cite{guo2020pct}  & xyz  & 16$\times$1024 & 93.2 & 92.4ms\\
PVT (Ours) & xyz & 16$\times$1024  & 93.7 & \textbf{45.5ms}\\
PVT (Ours) & xyz,nor & 16$\times$1024  & \textbf{94.1} & 48.6ms\\
\hline

\end{tabular}  
\end{center}  
\caption{Results on ModelNet40 \cite{20153D}. Compared with previous Transformer-based models, our PVT achieves the state-of-the-art accuracy with 10$\times$ measured speed-up on average.} 
\label{cla}
\end{table} 

\begin{table}  
\centering
\begin{center} 
\begin{tabular}{|c|cccc|}   
\hline
Input & PT$^1$ & PT$^2$ & PCT & Ours\\
\hline
16$\times$512 & 247.5 & 430.7 & 76.4 & 31.5  \\
16$\times$1024 & 320.6 & 530.2 &92.4& 45.5\\
8$\times$2048 & 460.5 & 720.4 &145.4& 71.2  \\
\hline
\end{tabular}  
\end{center}  
\caption{The speed comparison of different models when the number of input points varies.} 
\label{speedcompa}
\end{table} 

\textbf{Data. }We conduct the classification experiment on the ModelNet40 \cite{modelnet40} dataset. ModelNet40 includes 12,311 CAD models from 40 different object classes, in which 9843 models are used for training and the rest for testing. We follow the experimental configuration of PointNet, i.e., for each model, we uniformly sample 1024 points with 3 channels (or 6) of spatial location (and normal) as input; the point cloud is re-scaled to fit the unit sphere.

\textbf{Setting.} We utilize random translation, random anisotropic scaling and random input dropout strategies to augment the input points data during training. During testing, no data augmentation or voting methods were used. For classification on ModelNet40, the SGD optimizer was used for 200 epochs with the batch size 32. We set the initial learning rate to 0.01 and adopt a cosine annealing schedule to adjust the learning rate at every epoch.

\begin{figure}
    \centering
     \includegraphics[width=\linewidth]{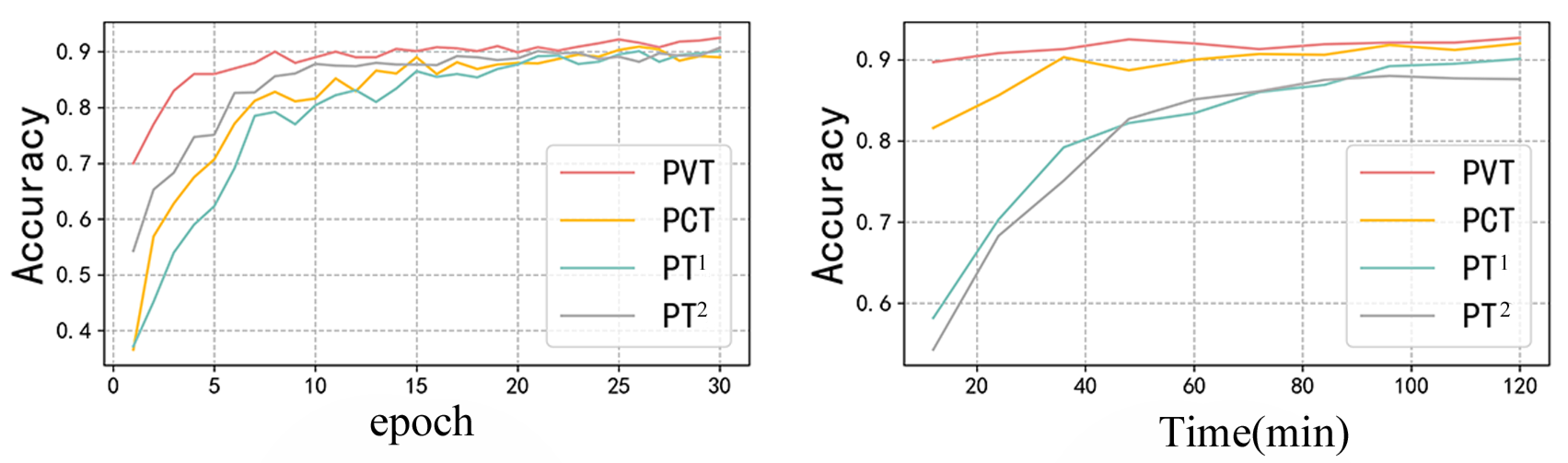}
    \caption{Accuracy of different Transformer-based
methods over time and epochs on ModelNet40. Our method performs the best in both equal-time and equal-epoch comparisons. The accuracy of 90\% can be achieved not only after 8 epochs but also in the shortest training time.}
    \label{epochandtime}
\end{figure}
\textbf{Results.} The results are shown in Tables \ref{cla} and \ref{speedcompa}, we can see that our PVT outperforms most previous models. Compared with its baselines, such as PCT, PT$^1$ and PT$^2$, PointASNL, our PVT not only achieves state-of-the-art accuracy of 94.1\%, but also has the best speed-accuracy trade-off (10$\times$ faster on average). Figure \ref{epochandtime} also provides an accuracy plot under equal-epoch setting. As can be seen, our method outperforms all Transformer-based methods, being \emph{the fastest} and \emph{most accurate} towards convergence.

%We also design narrower versions of PVT by reducing the number of resolutions to 50\% (denoted as 0.5×R). The resulting model requires only 53.5\% latency of PointNet, and it still outperforms several \emph{point}-based methods with sophisticated neighborhood aggregation including PCT, PointNet++ and DGCNN, which are almost an order of magnitude slower.
\begin{table}  
\centering
\begin{center} 
\begin{tabular}{|l|cccc|}   
\hline
Model & Params & FLOPs & SDA($\%$) & OA($\%$)\\
\hline
PointNet & 3.47M & 0.45G & 0.0 & 89.2  \\
PointNet++(SSG) & 1.48M & 1.68G &43.5& 90.7\\
PointNet++(MSG) & 1.74M & 4.09G &47.6& 91.9  \\
DGCNN  & 1.81M & 2.43G &57.2& 92.9 \\
\hline
PointASNL  & 3.98M & 5.92G &39.8& 93.1 \\
PT$^1$  & 21.1M & 5.05G &32.5& 92.8\\
PT$^2$  & 9.14M & 17.1G &65.4 & 93.7\\
PCT & 2.88M & 2.17G&24.6& 93.2\\
PVT(Ours) & \textbf{2.76M}& \textbf{1.93G} & \textbf{9.1} &  \textbf{94.1}\\
\hline
\end{tabular}  
\end{center}  
\caption{Computational resource requirements. SDA means the rate of total runtime on structuring the sparse data.}
\label{computational}
\end{table} 
\subsection{Analysis of computational requirements}
Now, we analyze the computational requirements of PVT and several other baselines by comparing the floating point operations required (FLOPs) and number of parameters (Params) in Table \ref{computational}. PVT has the lowest memory requirements with only 2.45M parameters, and also puts a low load on the processor of only 1.62G FLOPs, yet delivers highly accurate results of 94.1\%. These characteristics
make it suitable for deployment on a mobile device. In addition, PT$^2$ has attained top accuracy on various point cloud learning benchmarks, but its shortcomings of slow inference time and high computing cost are also obvious.
In the summary in Table \ref{computational}, PVT only spends 6.3\% of the total runtime on structuring the irregular data, which is much lower than previous Transformer-based models. 
Overall, PVT has the best accuracy and the lowest computational and memory requirements compared with its baselines. 
\subsection{Object part segmentation}
\begin{figure*}
    \centering
     \includegraphics[width=\linewidth]{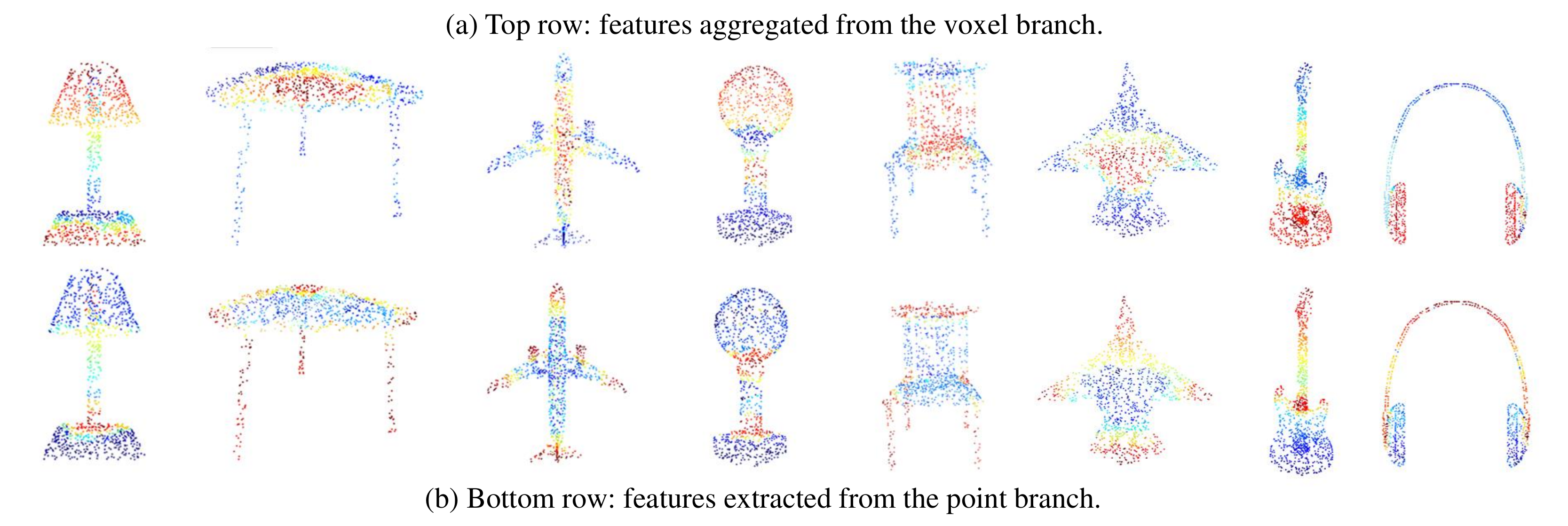}
    \caption{We demonstrate the output features extracted from two branches using Open3D \cite{open3d}. The \emph{voxel} branch focuses on the large, continuous parts, while the \emph{point}-based captures the global shape details.}
    \label{fig:disentangle}
\end{figure*}

\begin{table*}[tb]  
\centering
\begin{center}  
\resizebox{\textwidth}{23mm}{
\begin{tabular}{|p{1.6cm}|p{0.5cm}p{0.5cm}p{0.5cm}p{0.5cm}p{0.5cm}p{0.5cm}p{0.5cm}p{0.5cm}p{0.5cm}p{0.5cm}p{0.5cm}p{0.5cm}p{0.5cm}p{0.5cm}p{0.5cm}p{0.5cm}p{0.5cm}|}  
\hline
\footnotesize Model & \footnotesize pIoU & \footnotesize Areo & \footnotesize Bag & \footnotesize Cap & \footnotesize Car & \footnotesize Chair & \footnotesize Ear  \footnotesize Phone & \footnotesize Guitar & \footnotesize Knife & \footnotesize Lamp &  \footnotesize Laptop &  \footnotesize Motor & \footnotesize Mug & \footnotesize Pistol & \footnotesize Rocket & \footnotesize Skate \newline \footnotesize Board & \footnotesize Table\\  \hline
\footnotesize \# Shapes & $ $ & \footnotesize2690 & \footnotesize76 &\footnotesize 55 & \footnotesize898 & \footnotesize3758 & \footnotesize69 & \footnotesize787 &\footnotesize392 & \footnotesize1547 & \footnotesize451 & \footnotesize202 & \footnotesize184 & \footnotesize283 & \footnotesize66 & \footnotesize152 & \footnotesize5271 \\ \hline
\footnotesize PointNet &83.7& 83.4& 78.7 &82.5& 74.9& 89.6 &73.0& 91.5& 85.9 &80.8& 95.3& 65.2& 93.0 &81.2 &57.9& 72.8 &80.6\\
\footnotesize P2Sequence & 85.1 & 82.6 & 81.8 & 87.5 & 77.3 & 90.8 &77.1 & 91.1 & 86.9 & 83.9 & 95.7 & 70.8 & 94.6 & 79.3 & 58.1 & 75.2 & 82.8\\
\footnotesize PointASNL & 86.1 & 84.1 & 84.7 & 87.9 & 79.7 & 92.2 & 73.7 & 91.0 & 87.2 & 84.2 & 95.8 & 74.4 & 95.2 & 81.0 & 63.0 & 76.3 & 83.2\\
\footnotesize RS-CNN & 86.2 & 83.5 & \textbf{84.8} & 88.8 & 79.6 & 91.2& \textbf{81.1} &\textbf{91.6} & 88.4 & 86.0 & \textbf{96.1} & 73.7 & 94.1& 83.4 & 60.5 & \textbf{77.7} & \textbf{83.6}\\
\footnotesize PT$^1$ & 85.9 & $-$ & $-$ & $-$ &$-$ &  $-$& $-$ & $-$ & $-$ & $-$ & $-$ & $-$ & $-$ &$-$  & $-$ & $-$ &$-$ \\
\footnotesize PCT & 86.4 & 85.0 &82.4 & \textbf{89.0} & 81.2& 91.9 & 71.5 &91.3 &88.1  & \textbf{86.3} & 95.8 &64.6  & \textbf{95.8} &  83.6&62.2  &77.6  &73.7 \\
\footnotesize PVT (Ours) & \textbf{86.6} & \textbf{85.3} & 82.1 & 88.7& \textbf{82.1} & \textbf{92.4} &75.5  & 91.0 & \textbf{88.9} & 85.6&95.4  &\textbf{76.2}  & 94.7 &\textbf{84.2}& \textbf{65.0} & 75.3 & 81.7\\
\hline
\end{tabular}  }
\end{center}  
\caption{Results of part segmentation on ShapeNet. pIoU means part-average Intersection-over-Union.}  
\label{part_seg}
\end{table*} 

\begin{table*}[tb]  
\centering
\begin{center}  
\resizebox{\textwidth}{20mm}{
\begin{tabular}{|l|cccccccccccccc|}  
\hline
\footnotesize Model & \footnotesize mIoU & \footnotesize Ceiling & \footnotesize Floor & \footnotesize Wall & \footnotesize Bean & \footnotesize Column & \footnotesize Window & \footnotesize Door & \footnotesize Chair & \footnotesize Table & \footnotesize Bookcase & \footnotesize Sofa & \footnotesize Board & \footnotesize Clutter\\  \hline
PointNet & 41.09 & 88.80 & 97.33 & 69.80 & 0.05 & 3.92 & 46.26 & 10.76 & 52.61 & 58.93 & 40.28 & 5.85 & 26.38 & 33.22     \\   
PointNet++ & 50.04 & 90.79 & 96.45 & 74.12 & 0.02 & 5.77 & 43.59 & 25.39 & 69.22 & 76.94 & 21.45 & 55.61 & 49.34 & 41.88 \\ 
PointNet++-CE & 51.56 & 92.28 & 96.87 & 74.77 & 0.02 & 7.04 & 46.78 & 25.42 & 69.13 & 79.18 & 26.67& 53.39 & 54.61& 44.03 \\ 
DGCNN  & 47.08 & \textbf{92.42} & 97.46 & 76.03 & \textbf{0.37} & 12.00 & 51.59 & 27.01 & 64.85 & 68.58 & 7.67 & 43.76 & 29.44 & 40.83\\
PVCNN  & 56.12 & 91.23 & 97.54 & 77.13 & 0.29 & 13.02 & 51.72 & 26.74 & 68.52 & 75.48 & 28.64 & 53.29 & 27.21 & 41.92\\
BPM \cite{2021Boundary} & 61.43 & $-$ & $-$ & $-$ & $-$ & $-$ & $-$ & $-$ & $-$ & $-$ & $-$ & $-$ & $-$ & $-$ \\ 
PointASNL \cite{2021Boundary} & 62.60 & 94.31 & 98.42 & 79.13 & 0.00 & 26.71 & 55.21 & 66.21 & 83.32 & 86.83 & 47.64 & 68.32 & 56.41 & 52.12 \\ 
IAF-Net \cite{IAF} & 64.60 & 91.41 & 98.60 & 81.80 & 0.00 & \textbf{34.90} & \textbf{62.00} & 54.70 & 79.70 & 86.90 & \textbf{49.90} & 72.40 & 74.80 & 52.10 \\ 
PVT(Ours)  & \textbf{68.21} & 91.18 &\textbf{98.76}& \textbf{86.23}& 0.31 & 34.21& 49.90 &\textbf{61.45} &\textbf{81.62}& \textbf{89.85} &48.20& \textbf{79.96} &\textbf{76.45}& \textbf{54.67}\\
\hline
\end{tabular} 
}
\end{center}  
\caption{Indoor scene segmentation results on the S3DIS dataset, evaluated on Area5. From this table, we can see that the proposed PVT outperforms most of previous 3D models in some categories signiﬁcantly, }  
\label{Area5}
\end{table*} 

\begin{table*}[tb]  
\centering
\begin{center}  
\resizebox{\textwidth}{24mm}{
\begin{tabular}{|l|ccccccccccccccccccccc|}  
\hline
\footnotesize Model & \footnotesize Latency & \footnotesize mIoU & \footnotesize \rotatebox{90}{Car} & \footnotesize \rotatebox{90}{Bicycle} & \footnotesize \rotatebox{90}{MotorCycle} & \footnotesize \rotatebox{90}{Truck} & \footnotesize \rotatebox{90}{Other-vehicle} & \footnotesize \rotatebox{90}{Person} & \footnotesize \rotatebox{90}{Bicyclist} & \footnotesize \rotatebox{90}{Motorcyclist} & \footnotesize \rotatebox{90}{Road} & \footnotesize \rotatebox{90}{Parking} & \footnotesize \rotatebox{90}{Sidewalk} & \footnotesize \rotatebox{90}{Other-ground} & \footnotesize \rotatebox{90}{Building} & \footnotesize \rotatebox{90}{Fence} & \footnotesize \rotatebox{90}{vegetation} & \footnotesize \rotatebox{90}{Trunk} & \footnotesize \rotatebox{90}{Terrain} & \footnotesize \rotatebox{90}{Pole}  & \footnotesize \rotatebox{90}{Traffic-sign}\\  \hline
PointNet$^{*}$ & 500 & 14.6 & 46.3 & 1.3 & 0.3 & 0.1 & 0.8 & 0.2 & 0.2 & 0.0 & 61.6 & 15.8 & 35.7 & 1.4 &41.4 &12.9 &31.0 &4.6 &17.6&2.4 &3.7    \\   
PointNet++$^{*}$ & 5900 & 20.1 & 53.7 & 1.9 & 0.2 & 0.9 & 0.2 & 0.9 & 1.0 & 0.0 & 72.0 & 18.7 & 41.8 & 5.6 &62.3 &16.9 &46.5 &13.8 &30.0 &6.2  &8.9  \\ 
PVCNN$^{*}$ & 146 & 39.0 & $-$ & $-$ & $-$ & $-$ & $-$ & $-$ & $-$ & $-$ & $-$ & $-$ & $-$ & $-$ & $-$& $-$ & $-$ & $-$ & $-$ & $-$ & $-$\\ 
TangentConv$^{*}$ & 3000 & 40.9 & 83.9 &63.9& 33.4& 15.4& 83.4& 90.8 &15.2& 2.7& 16.5& 12.1& 79.5& 49.3 &58.1& 23.0 &28.4& 8.1& 49.0 &35.8 &28.5 \\ 
PointASNL & 7260 & 46.8& 87.4& 74.3 &24.3 &1.8& 83.1& 87.9& 39.0 &0.0 &25.1& 29.2& \textbf{84.1}& 52.2& 70.6& 34.2 &57.6 &0.0 &43.9 &57.8& 36.9 \\ 
RandLA-Net$^{*}$ & 880 & 53.9 & 94.2 &26.0 &25.8& \textbf{40.1}& 38.9& 49.2& 48.2& 7.2& \textbf{90.7}& 60.3 &73.7& 20.4 &86.9& 56.3 &81.4& 61.3& 66.8 &49.2 &47.7\\
KPConv$^{*}$ & 3560 & 58.8 & 96.0& 30.2& 42.5 &33.4 &44.3 &61.5 &\textbf{61.6}& 11.8& 88.8 &61.3& 72.7& 31.6 &\textbf{90.5} &\textbf{64.2} &\textbf{84.8} &69.2 &69.1& \textbf{56.4}& 47.4\\
BAAF & 6880 & 59.9& 90.9& 74.4 &\textbf{62.2}& 23.6& 89.8& \textbf{95.4} &48.7 &31.8& 35.5 &46.7& 82.7& 63.4 &67.9& 49.5 &55.7& 53.0& 60.8& 53.7& 52.0\\ 
SPVCNN $^{*}$& \textbf{110} & 63.7 & $-$ & $-$ & $-$ & $-$ & $-$ & $-$ & $-$ & $-$ & $-$ & $-$ & $-$ & $-$ & $-$& $-$ & $-$ & $-$ & $-$ & $-$ & $-$\\ 
PVT(Ours)  & 142 & \textbf{64.9} &\textbf{97.7} & \textbf{76.9} &59.8& 30.5& \textbf{91.8}& 94.4 &49.2 &\textbf{34.3}& 42.3 &\textbf{62.1}&81.3& \textbf{65.6} &70.9& 47.5 &68.7& \textbf{86.2}& \textbf{70.8}& 52.1& \textbf{54.0}\\

\hline
\end{tabular} 
}
\end{center}  
\caption{Results of outdoor scene segmentation on SemanticKITTI. $*$: results directly taken from \cite{tang2020searching}}  
\label{KITTI}
\end{table*} 
\textbf{Data. }The model is also evaluated on ShapeNet Parts \cite{shapenet}. It is composed of a total of 16,880 models (14,006 models are used for training, the rest for testing), each of which has 2 to 6 parts and the whole dataset is labeled in 50 different parts. A total 2048 points are sampled from each model as input and only few point sets have six labeled parts. We directly adopt the same train-test split as PointNet in our experiment.

\textbf{Setting. } 
The same training setting as in our classification task was adopted. For this task on ShapeNet Part, we train our model using the SGD optimizer for 200 epochs with the batch size 16. The initial learning rate was set to 0.1, with a cosine annealing schedule to adjust the learning rate at every epoch.

\textbf{Results.} Table \ref{part_seg} lists the class-wise segmentation results. Part-average IoU is used to evaluate our model, which is given both overall and for each object category. The results show that our PVT makes an improvement of 2.9\% over PointNet. Compared with all baselines, PVT attains the top pIoU with 86.6\%. 

\textbf{Visualization.}  In Figure \ref{fig:disentangle}, we illustrate output features from the \emph{point} and \emph{voxel} branches respectively, where a warmer color represents larger magnitude. As we can see, the \emph{voxel} branch captures large, continuous parts while the \emph{point}-based counterpart captures global shape details (e.g., table legs, airplane wings, and tail).

\begin{table}  
\centering
\begin{center} 
\begin{tabular}{|l|ccc|}   
\hline
Model & mIoU(\%) & OA(\%) & mAcc(\%)\\ 
\hline
PointNet & 47.6 & 78.5 & 66.2   \\
G+RCU \cite{2017Exploring} & 49.7 & 81.1 & $-$\\
TangentConv \cite{2018Tangent} & 52.8 & $-$ & $-$    \\
RSNet \cite{mehta2018rs} & 56.5 & $-$ & 66.5 \\
3P-RNN \cite{ye20183d} & 56.3 & 86.9 & $-$ \\
SGPN \cite{wang2018sgpn} & 50.4 & $-$ & $-$\\
SPGraph \cite{2018Large} & 62.1 & 85.5 & 73.0\\
HAPGN \cite{9141427} & 62.9 & 85.8 & $-$\\
PVCNN & 63.2 & 85.8 & 72.5\\
ShellNet \cite{2019ShellNet} & 66.8 & 87.1 &$-$\\
PVT(Ours) & \textbf{69.2}& \textbf{88.3} &  \textbf{76.2}\\
\hline
\end{tabular}  
\end{center}  
\caption{Results of semantic segmentation of 3D indoor scenes on S3DIS, evaluated with 6-fold cross-validation. From this table, we can see that the proposed PVT outperforms most of previous 3D models.} 
\label{6fold}
\end{table} 

\subsection{Indoor Scene Segmentation}
\textbf{Data.} To further assess our network, we conduct semantic segmentation task on S3DIS dataset \cite{DBLP:journals/corr/ArmeniSZS17}, which includes 273 million points from six indoor areas of three different buildings. Each point is annotated with a semantic label from 13 classes—e.g., beam, bookcase, chair, column, and window. We follow \cite{2017SEGCloud} and \cite{qi2017pointnet} and uniformly sampled points from blocks of area size $1m\times 1m$, where each point is represented by a 9D vector (XYZ, RGB, and normalized spatial coordinates).

\textbf{Setting. }
During the training process, we generate training data by randomly sampling
4,096 points from each block on-the-fly. Following \cite{2018Dynamic}, we utilize Area-5
as the test scene and all the other areas for training. Note that the position and RGB information of points are used to as the input features. The same training setting as in our classification task is adopted, but requires 50 epochs with the batch size 8 to fulfil the experiment.
\begin{figure*}
    \centering
    \includegraphics[width=\linewidth]{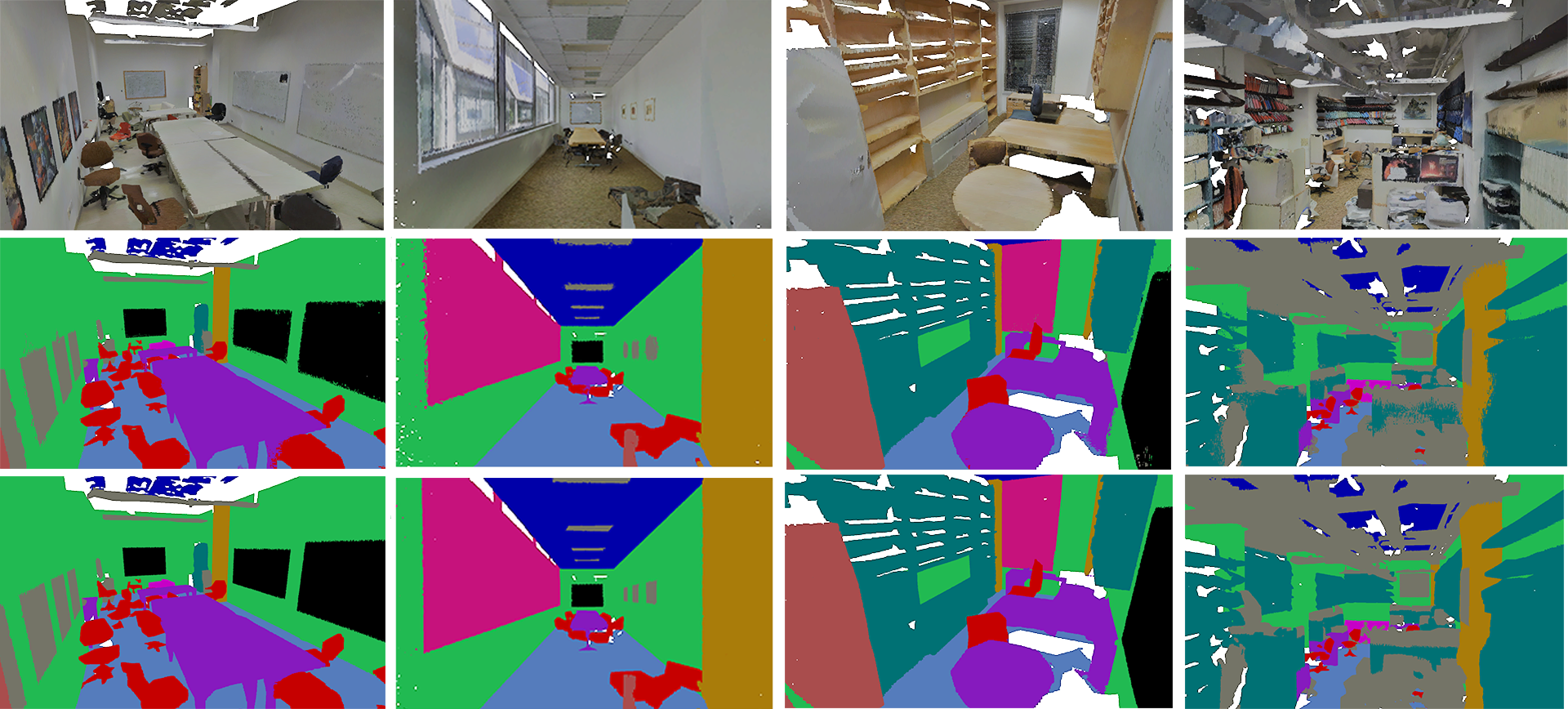}
    \caption{Visualization of semantic segmentation results on the S3DIS dataset. The input is in the top row, PVT predictions on the middle, the ground truth on the bottom.}
    \label{fig:vis}
\end{figure*}

\textbf{Results.} The results are presented in Tables \ref{Area5} and Table \ref{6fold}. From Tables \ref{Area5}, we can see that our PVT attains mIoU of 67.3\%, which outperforms MLPs-based frameworks such as PointNet \cite{qi2017pointnet} and PointNet++ \cite{qi2017pointnet++}, graph-based methods such as DGCNN \cite{2018Dynamic}, attention-based models such as PointASNL \cite{2020PointASNL}. Moreover, PVT attains mIoU of 69.2\% under 6-fold cross-validation, substantially outperforming most prior models.

\textbf{Visualization.} Fig.\ref{fig:vis} shows the visualization of PVT’s predictions. The predictions of our network are very close to the ground truth. PVT captures detailed semantic structure in complex 3D scenes, which is important in our network.

\subsection{Outdoor semantic Segmentation}
\textbf{Data.} To further evaluate our network, we conduct outdoor semantic segmentation task on SemanticKITTI \cite{semanticKT} dataset, which includes 43,552 densely labeled LIDAR scans belonging to 21 sequences. Each scan is a large-scale point cloud with $\sim10^5$ points and spanning up to 160×160×20 m in 3D space. Officially, the sequences $00\sim07$ and $09\sim10$ (19,130 scans) are used for training, the sequence 08 (4071 scans) for validation, and the sequences $11\sim21$ (20,351 scans) for online testing. The raw 3D points only have 3D coordinates without color information. 

\textbf{Setting. }Based on UNet, we build our backbone network for large-scale point clouds segmentation with a stem, four down-sampling and four up-sampling stages, and the dimensions of these nine stages are 32, 64, 64, 128, 256, 256, 128, 64, and 64, respectively.
As for the voxel branch, the voxel resolution is 0.05 m for segmentation experiments.
We train PVT for 100 epochs on a single GeForce RTX 3090 GPU with the batch size 8. In addition, the Adam optimizer is employed to minimize the overall loss;
the learning rate starts from 0.01 and decays with a rate of 0.5 after every 10 epochs.

\textbf{Results.} As shown in Table \ref{KITTI}, PVT outperforms the previous state-of-the-art point-based model BAAF by 5.0\% in mIoU with 48× measured speedup. Compared with strong attention-based PointASNL and point-based KPConv, our PVT achieves +18.1\% and +6.1\% mIoU improvements, with 51× and 25× measured speedup respectively.

\section{Ablation Studies}
In this section, we conduct extensive ablation study to analyze the effectiveness of different components of PVT block. The results of the ablation study are summarized in Tables \ref{ablation} and \ref{ablation:MS}.

\begin{table}  
\centering
\begin{center} 
\begin{tabular}{|c|cccccc|}   
\hline
Model & PB & VB & shifting & NPB & OA(\%) & Latency\\ \hline
A  &\XSolidBrush & \Checkmark & \Checkmark& 3 & 92.8 & 33.5\\
B   & \Checkmark & \XSolidBrush & \Checkmark & 3 & 92.3 & 25.2\\
C    & \Checkmark & \Checkmark & \XSolidBrush & 3 & 93.0 & 47.9\\ 
D  & \Checkmark & \Checkmark & \Checkmark & 2 & 93.1 & 36.4\\
E   & \Checkmark & \Checkmark & \Checkmark & 4 & 93.6 & 72.5\\ 
PVT $^{full}$ & \Checkmark & \Checkmark & \Checkmark & 3 & \textbf{94.1} & 48.6 \\
\hline
\end{tabular}  
\end{center}  
\caption{Ablation study on ModelNet40. PB and VB mean the point-based branch and the voxel-based branch; NPB denotes the number of PVT blocks; shifting means all self-attention modules
adopt the cyclic shifted box partitioning method. } 
\label{ablation}
\end{table} 

\begin{table}[tb]
\centering
\begin{center}  
\begin{tabular}{|c|cc|}
\hline
Ablation  & ModelNet40(OA) & ShapeNet(mIoU)  \\
\hline
MLP  & 92.6           & 85.3                \\
EdgeConv & 93.1          & 85.8          \\
w/o rel. pos  & 93.2       & 86.3    \\
PVT$^{full}$  & \textbf{94.1} & \textbf{86.6}\\
\hline
\end{tabular}
\end{center}  
\caption{Ablation study on the relative-attention and relative position bias on two benchmarks. MLP: replace relative-attention with MLP layer in our architecture. EdgeConv: replace relative-attention with EdgeConv layer in our architecture. no rel. pos: the relative attention without an additional relative position bias term (see Eq \ref{bias}).}
\label{ablation:MS}
\end{table}

\textbf{Impact of the voxel-based and point-based branches.} We set two baselines: A and B. Model A only encodes global context features by the \emph{point} branch, and Model B only encodes local features by the \emph{voxel} one. As reported in Table \ref{ablation}, the Baseline model A obtains a low accuracy of 92.8\% on classification benchmarks, and model B gets 92.3\%. When we combine local and global features (PVT$^{full}$), there is a notable improvement in both tasks. This means our network can take advantage of the combination of two branches, which provide richer information about the points.

\textbf{Effect of the cyclic shifted windows scheme.} Ablation of the \emph{shifted window} method on classification is reported in Table \ref{ablation} (Model C). The network with the \emph{shifted} window partitioning (PVT$^{full}$) outperforms the Model C without shifting at each layer by +1.0\% OA on ModelNet40. The results indicate the effectiveness and efficiency of using cyclic shifted window to build cross-window information interaction in the preceding layers.

\textbf{Impact of the number of PVT blocks. } we validate the impact of the PVT block by controlling the number of PVT blocks and report the results in Table \ref{ablation}. From this table, we can conclude the following: on one hand, reducing the number of PVT blocks can save latency, for example, compared with PVT$^{full}$, Model D saves 25\% latency but incurs a loss on accuracy; on the other hand, increasing the number of PVT blocks from PVT$^{full}$ can hardly support Model E accuracy benefit but leads to an increase on latency. To balance between speed and accuracy, we adopt 3 PVT blocks as our full model.

\textbf{Effect of Relative-attention.} We validate the impact of RA module used in our network. The results are shown in Table \ref{ablation:MS}. We set two baselines. "MLP" is a no-attention baseline that replaces RA with a MLP layer. "EdgeConv" is a more advanced no-attention baseline that replaces RA with a EdgeConv layer. EdgeConv performs feature aggregating at each point and enables each point to exchange information with its neighboring points, but does not leverage attention mechanisms. We can see that the RA module achieves better results than the no-attention baselines. The performance gap between RA and MLP baselines is significant: 94.1\% vs. 92.6\% and 86.6\% vs. 85.3\%, an respectivel improvement of 1.5 and 1.3 absolute percentage points. Compared with EdgeConv baseline, our RA module also achieves improvements of 0.9 and 0.9 absolute percentage points, respectively.

\textbf{Effect of relative position representations.} Finally, we also investigate the impact of RPR used in the RA module. As demonstrated in Table \ref{ablation:MS}, our PVT with RPR yields +0.8\% OA/+0.4\% mIoU on ModelNet40 and ShapeNet in relation to those without this term respectively, indicating the effectiveness of the RPR.

\section{Conclusion and Future Work}
In this work, we present PVT for efficient point cloud learning.
PVT is found to surpass previous Transformer-based and attention-based models in efficiency. It achieves this by deeply combining the advantages from both \emph{voxel}-based and \emph{point}-based networks. To reduce the computation cost, we design a GPU-based SWA computing method that has linear computational complexity with respect to voxel resolution and bypasses the invalid empty voxel computations. In addition, 
we study two different self-attention variants to gather fine-grained features about the global shape according to different scales of point clouds.

In the future, we expect to promote the primary structure for other research areas, such as point cloud pretraining, generation, and completion.

% \section*{Acknowledgments}
% This work was partially supported by the Zhejiang Provincial Natural Science Foundation of China (LGF21F20012), the National Natural Science Foundation
% of China (No.61602139), and the Graduate Scientific Research Foundation of Hangzhou Dianzi University (CXJJ2021082, CXJJ2021083).

{\small
\bibliographystyle{ieee}
\bibliography{egbib}
}

\end{document}